\documentclass[sigconf, screen, nonacm=true]{acmart}

\usepackage[english]{babel}
\usepackage[utf8]{inputenc} 
\usepackage[T1]{fontenc}    
\usepackage{hyperref}       
\usepackage{url}            
\usepackage{booktabs}       
\usepackage{multirow}
\usepackage{amsmath}  
\usepackage{amsfonts}       
\usepackage{nicefrac}       
\usepackage{microtype}      
\usepackage{xcolor}         
\usepackage{cleveref}
\usepackage{csquotes}
\usepackage[htt]{hyphenat}
\usepackage{graphicx}  
\usepackage{ulem}  
\usepackage{caption}
\usepackage{subcaption}
\usepackage{adjustbox}
\usepackage{xparse}
\usepackage{soul}        
\usepackage{siunitx} 

\title{Learned Hallucination Detection in Black-Box LLMs using Token-level Entropy Production Rate}

\author{Charles Moslonka}
\orcid{0000-0002-8719-1510}
\affiliation{%
	\institution{Artefact Research Center}
	\city{Paris}
	\country{France}
}
\affiliation{%
	\institution{MICS, CentraleSupélec, Université Paris-Saclay}
	\city{Gif-sur-Yvette}
	\country{France}
}
\email{{name.surname}@artefact.com}

\author{Hicham Randrianarivo}
\orcid{0009-0006-1764-5253}
\affiliation{%
	\institution{Artefact Research Center}
	\city{Paris}
	\country{France}
}
\affiliation{%
	\institution{MICS, CentraleSupélec, Université Paris-Saclay}
	\city{Gif-sur-Yvette}
	\country{France}
}
\email{{name.surname}@artefact.com}

\author{Arthur Garnier}
\affiliation{%
	\institution{Ardian}
	\city{Paris}
	\country{France}
}
\email{{name.surname}@ardian.com}

\author{Emmanuel Malherbe}
\orcid{0009-0006-0898-6873}
\affiliation{%
	\institution{Artefact Research Center}
	\city{Paris}
	\country{France}
}
\email{{name.surname}@artefact.com}

\begin{document}
\begin{abstract}
Hallucinations in Large Language Model (LLM) outputs for Question Answering (QA) tasks can critically undermine their real-world reliability. This paper introduces a methodology for robust, one-shot hallucination detection, specifically designed for scenarios with limited data access, such as interacting with black-box LLM APIs that typically expose only a few top candidate log-probabilities per token. Our approach derives uncertainty indicators directly from these readily available log-probabilities generated during non-greedy decoding. We first derive an Entropy Production Rate (EPR) that offers baseline performance, later augmented with supervised learning. Our learned model leverages the entropic contributions of the accessible top-ranked tokens within a single generated sequence, without multiple re-runs per query. Evaluated across diverse QA datasets and multiple LLMs, this estimator significantly improves token-level hallucination detection over state-of-the-art methods. Crucially, high performance is demonstrated using only the typically small set of available log-probabilities (e.g., top-10 per token), confirming its practical efficiency and suitability for API-constrained deployments. This work provides a lightweight technique to enhance the trustworthiness of LLM responses, at the token level, after a single generation pass, for QA and Retrieval-Augmented Generation (RAG) systems. Our experiments confirmed the performance of our method against existing approaches on public dataset as well as for a financial framework analyzing annual company reports.

\end{abstract}

\maketitle

\section{Introduction}

\begin{figure*}
	\centering
	\includegraphics[width=0.8\linewidth]{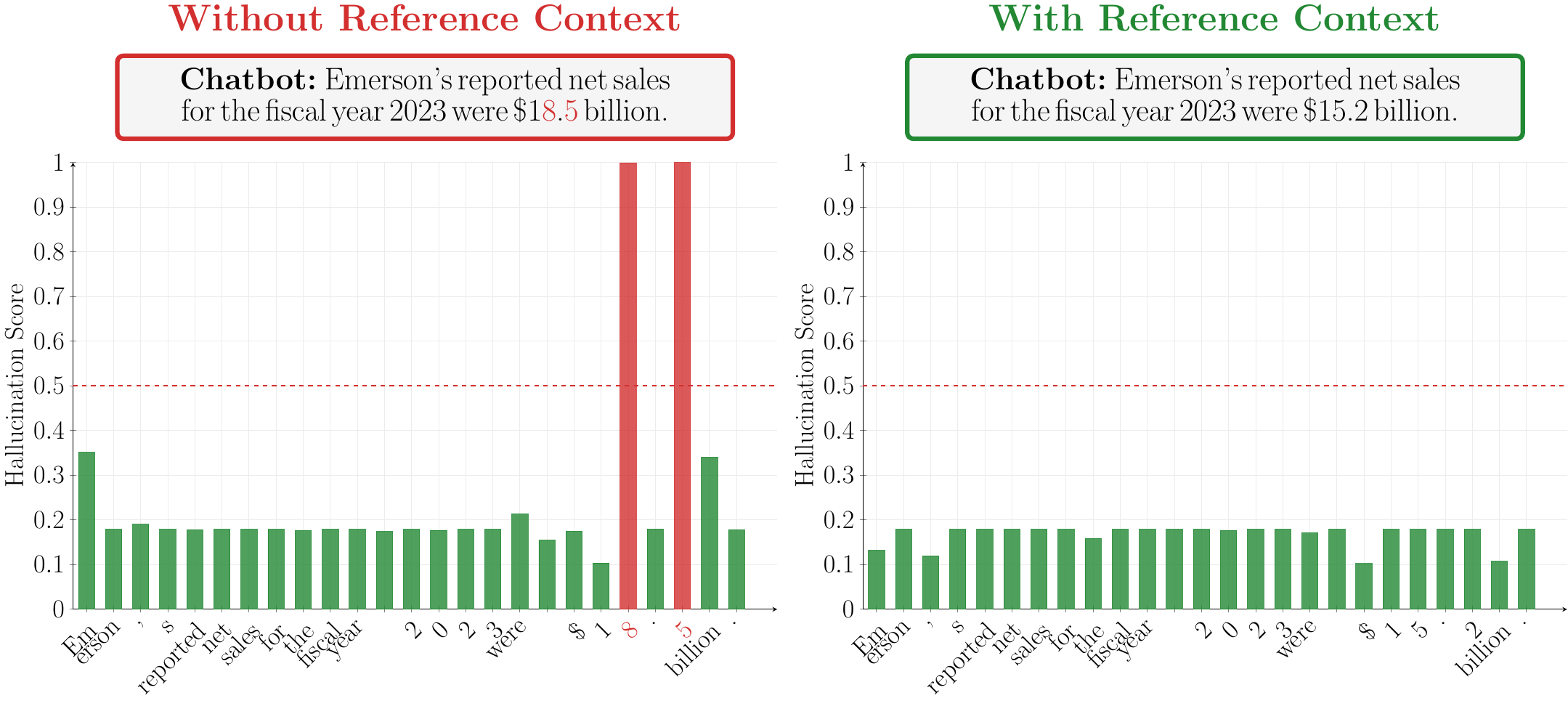}
	\caption{Illustration of token-level uncertainty detection in a financial RAG pipeline.}
    \Description{We show the token-level scores of a LLM asked what is Emerson's net sales in 2023, in two different cases. Without retrieved contents, the generated answer has low uncertainty on all tokens except on the digits of the answer (false answer). With retrieved context, all tokens have a low score, and the digits returned are correct.}
	\label{fig:overview}
\end{figure*}

Large Language Models (LLMs) have demonstrated remarkable capabilities across a wide range of tasks~\cite{petroniLanguageModelsKnowledge2019,weiEmergentAbilitiesLarge2022,openaiGPT4TechnicalReport2024}.
However, a significant and persistent challenge is their propensity to generate hallucinations~\cite{rohrbachObjectHallucinationImage2019,bakerHallucinatingFaces2000}.
These are defined as LLM-generated content that is factually incorrect, nonsensical, or unfaithful to a provided source, despite often appearing plausible and coherent~\cite{jiSurveyHallucinationNatural2023a,sriramananLLMCheckInvestigatingDetection}.
The impact of hallucinations can be severe, particularly in safety-critical applications such as medicine or infrastructure engineering~\cite{kimMedicalHallucinationsFoundation2025,liuExploringEvaluatingHallucinations2024}, leading to the propagation of misinformation and erosion of user trust.
Therefore, detecting and quantifying the uncertainty associated with LLM outputs is paramount to building trust and enabling responsible deployment.

The practical deployment of Uncertainty Quantification (UQ)~\cite{sullivanIntroductionUncertaintyQuantification2015} and hallucination detection methods is often constrained by limited access to LLM internals. These limitations are especially true when interacting with proprietary models through APIs, which may expose only a small number of top-$K$ log-probabilities per token (where $K$ might be around $15$\footnote{The maximum value is $K=20$ for the OpenAI API at the time of writing.}). This black-box setting restricts the applicability of techniques requiring access to full logits, hidden states, or architectural modification~\cite{manakulSelfCheckGPTZeroResourceBlackBox2023}. 
Moreover, many real-world applications require one-shot (or \enquote{single-round}~\cite{liuUncertaintyQuantificationConfidence2025}) detection—the ability to assess the reliability of a single generated sequence without the need for multiple, often costly, model inferences for the same input query to measure output variability.

This paper introduces a UQ methodology rooted in information theory, tailored for these black-box, one-shot scenarios. Our approach leverages the accessible log-probabilities to derive entropic measures of model hesitation during the generation process. We define and evaluate a metric termed the Entropy Production Rate (EPR) of a generated sequence. We show that EPR, calculated as the average entropy of the token probability distributions across the entire sequence, serves as an initial, unsupervised estimator of the degree of hesitation exhibited by the model during inference.

Our second contribution is a supervised learning model that adapts entropy measures to a dataset of annotations. By training on the entropic contributions of top-ranked tokens across the sequence, we learn an estimator that more accurately distinguishes between faithful and hallucinatory responses. This estimator can also highlight high-uncertainty tokens in a generated sequence, as displayed in Figure~\ref{fig:overview}.

We detail the entropic framework, its theoretical and practical underpinnings under limited log-probability access, and its empirical validation. We showcase its potential as a practical tool for improving the reliability of LLM systems, using RAG in a financial scenario as an example. To facilitate adoption and future research, we release our methods (EPR and WEPR) as an open-source Python package available at \url{https://github.com/artefactory/artefactual/}.

\section{Related Work}
\paragraph{UQ in Large Language Models.}
The primary goal of UQ in LLMs is to enable them to signal their confidence — to \enquote{know what they do not know}~\cite{kadavathLanguageModelsMostly2022,azariaInternalStateLLM2023}.
LLM uncertainty arises from traditional aleatoric (data-inherent) and epistemic (model-based) sources~\cite{gal2016uncertainty}, and additionally from their generative characteristics like divergent multi-step reasoning paths and decoding stochasticity~\cite{liuUncertaintyQuantificationConfidence2025}. Traditional neural network UQ approaches, such as training modifications for aleatoric uncertainty~\cite{hullermeier2021aleatoric} or dropout-based methods for epistemic uncertainty~\cite{postels2019sampling,gal2016dropout}, are often ill-suited or prohibitively costly for LLMs due to training expense or architectural differences. Developing UQ techniques tailored to LLM intricacies remains an active research area~\cite{quevedo2024detecting}.

Information-theoretic entropy has emerged as a cornerstone for assessing LLM uncertainty and hallucinations, based on the hypothesis that higher output distribution entropy often correlates with a greater likelihood of error~\cite{shannonMathematicalTheoryCommunication1948}.
Researchers have employed diverse entropy measures to capture different facets of LLM uncertainty~\cite{gabrieEntropyMutualInformation2018}. Predictive entropy~\cite{kweonUncertaintyQuantificationDecomposition2025} assesses overall output distribution uncertainty. Semantic Entropy~\cite{kuhnSemanticUncertaintyLinguistic2023} quantifies uncertainty over semantic meaning by clustering multiple generated responses, with Discrete Semantic Entropy~\cite{farquharDetectingHallucinationsLarge2024,cecereMonteCarloTemperature2025} using frequency-based cluster probabilities. 
Answer Entropy~\cite{liuEnhancingHallucinationDetection2025}, common in QA, measures variability in complete answer strings from multiple trials, sometimes enhanced with noise injection. Particularly relevant to our work, token-level entropy~\cite{kadavathLanguageModelsMostly2022} measures per-step uncertainty from the next-token probability distribution and has been used to predict errors in QA and mathematical reasoning~\cite{manakulSelfCheckGPTZeroResourceBlackBox2023,jooEntropybasedSentencelevelHallucination2025}.

\paragraph{Entropy-driven detection techniques and challenges.}
Direct application of entropy scores, often via thresholding, is a common approach for hallucination detection or calibration~\cite{manakulSelfCheckGPTZeroResourceBlackBox2023,kuhnSemanticUncertaintyLinguistic2023,kadavathLanguageModelsMostly2022}. More sophisticated techniques include Semantic Entropy Probes (SEPs)~\cite{kossenSemanticEntropyProbes2024}—learned models predicting Semantic Entropy from single-pass hidden states for efficiency—or Shifting Attention to Relevance (SAR)~\cite{duanShiftingAttentionRelevance2024}, which weights tokens by relevance. However, entropy-based methods face challenges like high-certainty hallucinations~\cite{simhiTrustMeIm2025}, where LLMs produce low-entropy incorrect outputs, necessitating strategies beyond simple thresholds. Moreover, the computational cost of multi-sample methods (e.g., for Answer or Semantic Entropy) can be prohibitive, highlighting the need for efficient single-shot techniques such as SEPs~\cite{kossenSemanticEntropyProbes2024}, Logit-induced Token Uncertainty (LogTokU)~\cite{maEstimatingLLMUncertainty2025}, and ours.

\section{Theoretical Framework for Entropic Uncertainty}\label{sec:theoretical_setup}

Our approach to estimate uncertainty in LLM generations is rooted in information theory~\cite{brillouinScienceInformationTheory2013,mezardInformationPhysicsComputation2008}. The core idea is to derive an entropic score reflecting the model's hesitation when generating tokens. Unlike methods requiring multiple model runs to assess output variability, our method is a one-shot analysis, leveraging the probabilistic information from a single generated sequence.
\subsection{Entropy computation during token generation.}
For a standard classification task with $N_c$ classes, given an input $q$, a model outputs probabilities $p_i = p(\text{class}_i|q)$ for each class $i \in \{1, \dots, N_c\}$. The uncertainty can be quantified by the Shannon entropy (in bits) of this distribution:
\begin{equation}
	H(q) = - \sum_{i=1}^{N_c} p_i \log_2(p_i) \geq 0.
\end{equation}
This entropy $H(q)$ is minimized to zero when the distribution is sharply peaked (high certainty) and maximized at $H_{\mathrm{max}}(q) = \log_2(N_c)$ for a uniform distribution (maximum uncertainty).

Extending this to LLM sequence generation, let $V = \{v_i\}_{i=1}^{|V|}$ denote the model's vocabulary, with size\footnote{e.g., $|V| = 2^{17} = 131{,}072$ for some contemporary models, such as \texttt{Mistral-Small-3.1-24B}~\cite{MistralSmall31}.} $|V|$. For a given input query $q$, the LLM generates a sequence of tokens $\mathcal{T} = \{t_1, t_2, \dots, t_{|\mathcal{T}|}\}$ ($\forall j, t_j \in V$). The generated sequence $\mathcal{T}$ depends on $q$, but we omit this dependency for clarity. At each generation step $j$, the LLM internally computes a probability distribution $\{p(v_i \mid q, t_{<j})\}_{i=1}^{|V|}$ for the next token $t_j$, conditioned on the preceding tokens $t_{<j} := \{t_1, \dots, t_{j-1}\}$ and the input $q$. The complete entropy for the $j$-th token's distribution is:
\begin{equation}
  H(q, t_{<j}) = - \sum_{i=1}^{|V|} p(v_i \mid q, t_{<j}) \log_2 p(v_i \mid q, t_{<j}).
\end{equation}
Note that this quantity does not depend on the sampled token $t_j$.

In black-box scenarios involving proprietary LLMs accessed via APIs, only the log-probabilities (or probabilities) for a small number, denoted by $K$, of top-ranked candidate tokens are exposed (e.g., $K \le 20$). Thus, we estimate the per-token entropy based on these top $K$ probabilities. Let $r: \{1,\dots,|V|\} \rightarrow \{1,\dots,|V|\}$ be the ranking operator on the vocabulary indices, meaning $r(k)$ is the index $i$ of the $k$-th token in decreasing order of $p(v_i \mid q,t_{<j})$.
The estimator can then be written as:
\begin{equation}\label{eq:H_K}
	\tilde{H}_K(q, t_{<j}) = - \sum_{k=1}^{K} p_{r(k),j} \log_2\left(p_{r(k),j}\right),
\end{equation}
where $p_{r(k),j} := p(v_{r(k)} | q, t_{<j})$ is the probability of the $k$-th ranked token at generation step $j$.

\subsection{Sufficiency of Top-$K$ Log-Probabilities for Entropy Estimation}
A critical question is whether $\tilde{H}_K(q, t_{<j})$, calculated from only the top $K$ probabilities, is a reliable proxy for the true entropy $H(q, t_{<j})$. The discrepancy arises from the missing tail of the distribution:
\begin{equation}
  \Delta H_K(q, t_{<j}) = H(q, t_{<j}) - \tilde{H}_K(q, t_{<j}) = - \sum_{k=K+1}^{|V|} p_{r(k),j} \log_2 p_{r(k),j}.
\end{equation}
We can establish an upper bound for this missing entropy by assuming the remaining probability mass $Q_K(q, t_{<j}) = 1 - \sum_{k=1}^{K} p_{r(k),j}$ is distributed uniformly among the remaining tokens considered sampleable. When the LLM generation uses a cutoff parameter $K_{\mathrm{samp}}\ll |V|$ (e.g., $K_{\text{samp}}=50$), probabilities beyond $K_{\text{samp}}$ are set to zero. In that case:
\begin{equation}
    \Delta H_{K, \mathrm{max}}(q, t_{<j}) = -Q_K(q, t_{<j}) \left[\log_2(Q_K(q, t_{<j})) - \log_2(K_{\text{samp}}\!-\!K)\right] \label{eq:delta_H_max_K_samp_tail}
\end{equation}

The quality of $\tilde{H}_K(q, t_{<j})$ as an estimator of the entropy within the relevant token pool (either full vocabulary or $K_{\text{samp}}$ limited) can be assessed by comparing $\tilde{H}_K(q, t_{<j})$ and $\Delta H_{K, \mathrm{max}}(q, t_{<j})$; a large ratio $\tilde{H}_K / \Delta H_{K, \mathrm{max}}$ suggests that $\tilde{H}_K(q, t_{<j})$ captures most of the relevant uncertainty.

\begin{figure*}
    \centering
    \includegraphics[width=\linewidth]{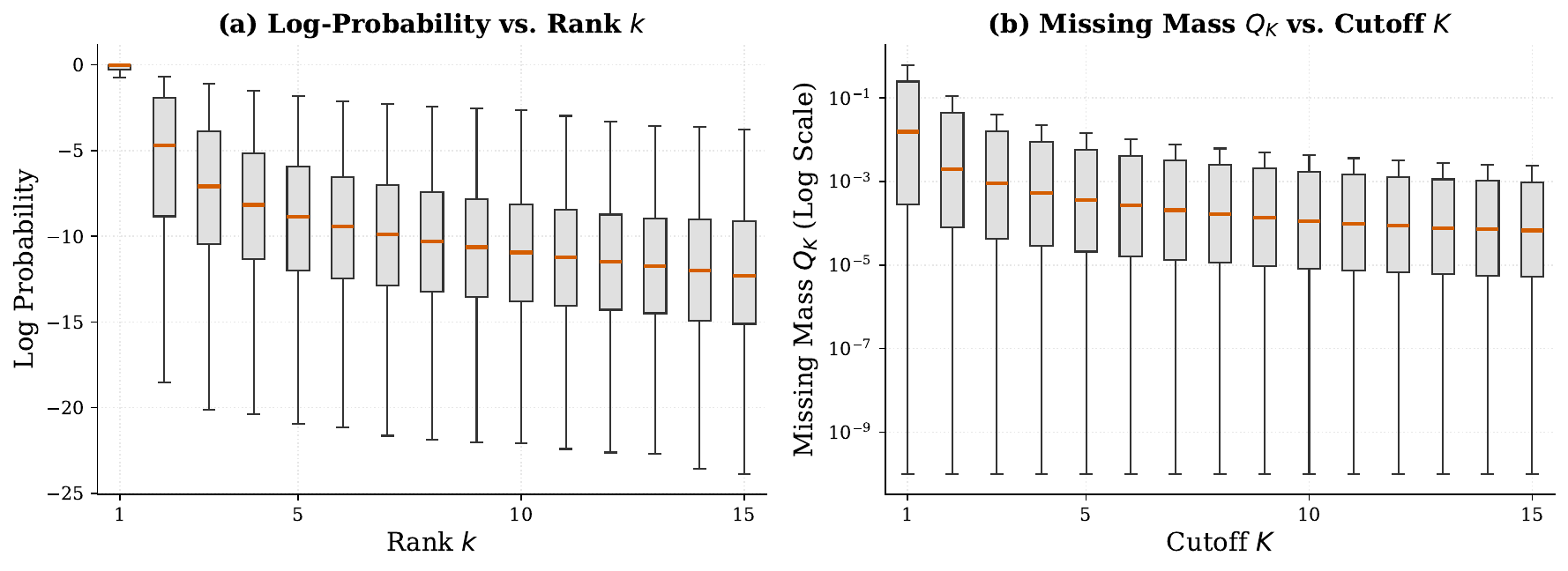}
    \caption{Analysis of probability mass concentration on $180$k tokens generated by \texttt{Mistral-Small-3.1-24B} ($T_{\text{samp}}=1.0$). \textbf{(a)} Distribution of log-probabilities for the token at rank $k$, showing a rapid decay in likelihood. \textbf{(b)} Distribution of the missing probability mass $Q_K$ (log scale) when truncating at cutoff $K$. The median $Q_K$ drops below $10^{-4}$ by $K=10$, validating the Top-$K$ approximation.}
    \Description{Two box-plots of token log-probability distributions.}
    \label{fig:logprob_concentration}
\end{figure*}

Empirically, we observe a concentration in the log-probabilities of lower-ranked tokens (e.g., for ranks $i \approx 10-15$), as illustrated in Figure~\ref{fig:logprob_concentration}.
Differences between successive $p_{r(k),j}$ tend to diminish for larger $k$ within the accessible $K$. When combined with $K_\mathrm{samp}$ sampling, this supports the idea that $\tilde{H}_K(q, t_{<j})$ captures the dominant portion of the uncertainty within the set of practically sampleable tokens, especially if $K$ is reasonably close to $K_{\text{samp}}$.

\begin{figure}[t]
\centering
\includegraphics[width=\linewidth]{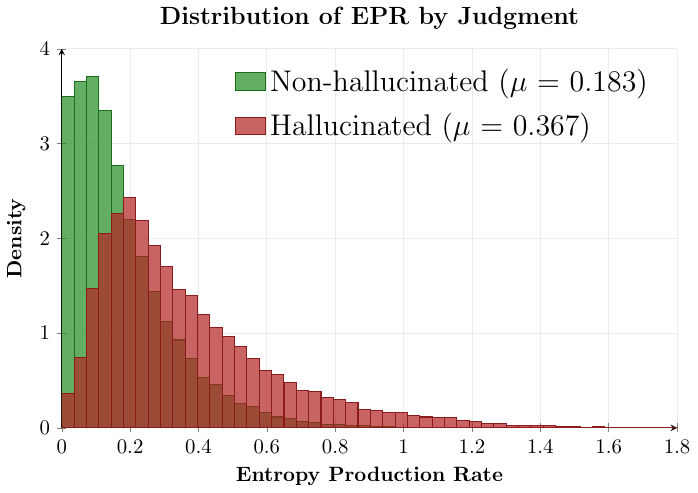}
\caption{Distributions of EPR scores on \texttt{Falcon-10B} responses. The separation between the \textit{Hallucinated} (red) and \textit{Non-hallucinated} (green) distributions illustrates the discriminative power of the metric.}
\Description{On one plot, two distributions of the EPR scores, colored differently according to their label/judgment, hallucinated or not. The mean EPR score is higher for hallucinated answers.}
\label{fig:histogram}
\end{figure}
\subsection{Influence of Sampling Temperature}
In practice, at $j$-th generation step, a LLM first predicts raw logits $l_{j,i}$ and converts them to probabilities $p_{j,i}$ by a softmax function with a temperature parameter $T_{\text{samp}}$, such that: $p_{j,i}(T_{\text{samp}}) = e^{l_{j,i}/T_{\text{samp}}} / \sum_{m=1}^{|V|} e^{l_{j,m}/T_{\text{samp}}}$. 
\newline
The per-token information entropy $H(q, t_{<j}; T_{\text{samp}})$ thus depends on $T_{\text{samp}}$, verifying:
\begin{itemize}
	\item As $T_{\text{samp}} \to 0^+$, $p_{j,i}(T_{\text{samp}})$ approaches $1$ for the token with the highest logit (assuming a unique maximum) and $0$ for others. Consequently, $H(q, t_{<j};  T_{\text{samp}}) \to 0$, reflecting high certainty.
	\item As $T_{\text{samp}} \to \infty$, $p_{j,i}(T_{\text{samp}})$ approaches $1/|V|$ for all tokens, leading to a uniform distribution. Therefore, the entropy $H(q, t_{<j}; T_{\text{samp}}) \to \log_2 |V|$, its maximum possible value, reflecting maximum uncertainty.
\end{itemize}
However, a critical distinction exists between the distribution used for \textit{sampling} tokens and the log-probabilities returned by modern inference engines. Systems like \texttt{vllm}, or the OpenAI API, typically expose log-probabilities computed from raw logits (equivalent to fixing $T=1$ for calculation), regardless of the $T_{\text{samp}}$ used to select the token $t_j$.
Consequently, while $T_{\text{samp}}$ dictates the trajectory of the generated sequence $\mathcal{T}$, our entropic features (defined in the following sections) measure the model's intrinsic hesitation based on its unscaled raw distribution. This decoupling ensures that our estimator remains mathematically well-defined and stable even if the user employs conservative decoding settings (e.g., low $T_{\text{samp}}$).

\subsection{Entropy Production Rate (EPR) of the sequence}
Based on the per-token entropy estimate $\tilde{H}_K(q, t_{<j})$, we define the Entropy Production Rate (EPR) for a generated sequence $\mathcal{T}$ of length $|\mathcal{T}|$ in response to query $q$. It is the average estimated entropy over the top $K$ accessible log-probabilities, across all tokens in the sequence:
\begin{align}
	\mathrm{EPR}_K(q, \mathcal{T}) &= \frac{1}{|\mathcal{T}|} \sum_{j=1}^{|\mathcal{T}|} \tilde{H}_K(q, t_{<j})\\ \nonumber  &= - \frac{1}{|\mathcal{T}|} \sum_{j=1}^{|\mathcal{T}|} \sum_{k=1}^{K} p_{r(k),j} \log_2(p_{r(k),j}).
	\label{eq:epr_definition}
\end{align}
This $\mathrm{EPR}_K$ may serve as a global measure of hesitation or uncertainty for the generated sequence, given the black-box constraints (see Fig.~\ref{fig:histogram})
In the following, we improve its alignment with ground-truth annotations via supervised learning.

\subsection{Supervision on Entropic Contributions by Rank}

We consider a dataset $\mathcal{D}$ where each entry is a query $q$, the generated sequence $\mathcal{T}$, and an annotation $Y \in \{0, 1\}$ indicating whether $\mathcal{T}$ is a correct (therefore considered non-hallucinatory) answer to $q$. Note that this annotation depends on the LLM used to generate $\mathcal{T}$. A realistic dataset size is on the order of hundreds to thousands of entries.

To build a more nuanced supervised detector, we define features based on the entropic contributions of tokens at specific ranks. For the $j$-th token generation and given $k \in \{1, \dots, K\}$, the entropic contribution of the $k$-th ranked token is 
$$s_{k,j} = -p_{r(k),j} \log_2(p_{r(k),j}).$$
Note that the actually sampled token at step $j$ may not be the highest-ranked token ($k=1$), especially with high $T_{\text{samp}}$. Our estimation leverages the probabilities of the tokens at fixed ranks $k$, regardless of which token was sampled.
We consider weighing each term of $\tilde{H}_K(q, t_{<j})$ (Eq.~\eqref{eq:H_K}) according to its rank $k$:

\begin{equation}
\label{S_beta}
	S_{\beta}(q, t_{<j}) = \beta_0 + \sum_{k=1}^{K} \beta_k s_{k,j},
\end{equation}

where $\beta \in \mathbb{R}^{K+1}$ are weights that aim to adapt the entropy measure and are learned using the annotations of $\mathcal{D}$, as described in the following. This token-wise quantity $S_{\beta}(q, t_{<j})$ is close to $\tilde{H}_K(q, t_{<j})$, but can be negative. 

We aggregate the token-wise scores to learn the parameters $\beta$ at the sequence level. A simple average of $S_\beta\left(q, t_{<j}\right)$ across the sequence captures the overall model hesitation. However, a single moment of high uncertainty (a "spike") can often be indicative of a hallucination, even if the rest of the sequence is generated confidently, and in such case $s_{k,j}$ values will be high for the uncertain generation, even at high ranks $k$. To capture these critical tokens, we also consider the maximum entropic contribution $s_{k,j}$ for each rank across the sequence.
This leads to our final scoring function, the Weighted Entropy Production Rate (WEPR):
\begin{equation}\label{eq:WEPR}
    \mathrm{WEPR}_{\beta,\gamma}(q,\mathcal{T}) = \frac{1}{|\mathcal{T}|}\sum_{j=1}^{|\mathcal{T}|} {S_{\beta}(q, t_{<j})} + \sum_{k=1}^K \gamma_k \max_j s_{k,j} 
\end{equation}
where $\gamma \in \mathbb{R}^{K}$
are parameters to be learned along with $\beta$ by maximizing on the annotations of the dataset $\mathcal{D}$:
\begin{align}
    \max_{\beta,\gamma} \sum_{\substack{q, \mathcal{T}, Y \in\mathcal{D}}} \! Y &\log \left[ \sigma \left(\mathrm{WEPR}_{\beta,\gamma} (q,\mathcal{T}) \right) \right]\\ \nonumber 
    +(1-Y) &\log \left[ \sigma\left(1 \!-\! \mathrm{WEPR}_{\beta,\gamma}(q,\mathcal{T})\right)\right]  
\end{align}
where $\sigma$ is the sigmoid function. This corresponds to a logistic regression loss, without regularization, on the averaged contributions $S_\beta(q, t_{<j})$ across tokens and the maximal $s_{k,j}$ values.
This way, the signal provided by the entropy along the sequence is aligned to the ground truth annotations.

Beyond this supervision, $\sigma\left(\mathrm{WEPR}_{\beta,\gamma}(q,\mathcal{T})\right) \in [0,1]$  
can be interpreted as a confidence score for the generated sequence $\mathcal{T}$ to be valid, with no hallucination. In particular, contrary to $\mathrm{EPR}_K(q)$, this quantity is scaled and can easily be displayed as a normalized score to the user.
Moreover, while the annotations are at the sequence level, we can also consider a token-level measure, computed on the entropic contributions features of each token:
\begin{equation}\label{eq:token_score}
\sigma\big(S_\beta\left(q, t_{<j}\right)\big) \in [0,1].
\end{equation}
This score can be interpreted as the localized risk of hallucination associated with the $j$-th token, allowing users to identify specific parts of a less reliable response. A significant advantage of this token-level score is its suitability for online estimation: it can be computed and displayed in real-time as the sequence is generated, token by token, without waiting for the full output $\mathcal{T}$.

\section{Experiments}\label{sec:experiments}
\subsection{Experimental Setup}

\subsubsection{Models and Datasets}\footnote{Code and data to reproduce the experiments presented in this paper are available at \url{https://github.com/artefactory/artefactual/releases/tag/ECIR2026}.}
We evaluated our methods on a suite of contemporary instruction-finetuned LLMs, including \texttt{Mistral-Small-3.1-24B-Instruct-2503}~\cite{MistralSmall31}, \texttt{Falcon3-10B-Instruct}~\cite{Falcon3}, \texttt{Phi-4}~\cite{abdinPhi4TechnicalReport2024}, and \texttt{Ministral-8B-Instruct-2410}~\cite{MinistralMinistrauxMistral}.
For Question Answering (QA) tasks, we primarily utilized the TriviaQA~\cite{joshiTriviaQALargeScale2017} (Wikipedia domain) and WebQuestions~\cite{berantSemanticParsingFreebase2013} datasets. We also explored the applicability of our entropic measures in a Retrieval-Augmented Generation (RAG)~\cite{lewisRetrievalAugmentedGenerationKnowledgeIntensive2021} context using a specialized dataset based on financial reports from the ArGiMi-Ardian Finance dataset~\cite{argimi2024Datasets} 
to observe how entropy varies with the provision of relevant context (we detail this particular application below). We focused on models generating relatively short answers, typical of non-extended reasoning QA.

\subsubsection{Answer Generation and Log-probability Extraction}
LLMs were served using the \texttt{vllm} inference library~\cite{kwon2023efficient}. To ensure variability and access to meaningful probability distributions for entropy calculation, we used non-greedy decoding with $T_{\text{samp}}=1.0$. The \texttt{top\_p} parameter was set to $1.0$, while the sampling cutoff was set to $K_{\text{samp}}=50$. This truncation allows us to compute the theoretical upper bound on missing entropy defined in Eq.~\eqref{eq:delta_H_max_K_samp_tail}. For each generated token, we extracted the accessible top-$K$ log-probabilities (typically $K \le 15$ or $K \le 20$ depending on the API/model).

\subsubsection{Hallucination Annotation Protocol}
Generated answers for the QA tasks were labeled as hallucinated or non-hallucinated (as displayed in Figure~\ref{fig:histogram}) using an LLM-as-a Judge~\cite{zhengJudgingLLMJudgeMTBench2023,wangLargeLanguageModels2023,adlakhaEvaluatingCorrectnessFaithfulness2024,rahmaniLLMJudgeLLMsRelevance2024} approach, employing \texttt{Gemma-3-12b-it}~\cite{teamGemma3Technical2025}. The judge LLM semantically compared the generated answer to the ground-truth answer (and any available aliases). If both answers were judged similar, the generated answer was labeled as non-hallucinated.
This automated annotation is scalable and suitable for one-shot analysis. However it may mislabel random correct guesses with high uncertainty or fail to flag confident but incorrect statements (high-certainty hallucinations).
To validate its reliability, we conducted a human evaluation. A group of 15 annotators, comprising Ph.D. and graduate students in Machine Learning from our laboratory, hand-labeled a sample set of $1{,}333$ output/ground-truth pairs.
The human-machine agreement was exceptionally high: they matched on $1{,}275$ instances ($95.7\%$), yielding a Cohen's Kappa of $\kappa = 0.898$ (see Table~\ref{tab:annotations}). This strong alignment confirms that the automated labels used for our large-scale training and evaluation are reliable proxies for human judgment.

\subsubsection{RAG Uncertainty Analysis}
Beyond general QA, we explored the applicability of our entropic methods for analyzing outputs in RAG pipelines. The premise is that uncertainty, as captured by EPR or WEPR, might increase when an LLM generates an answer without sufficient supporting context.
We demonstrated this capability using the \texttt{ArGiMi-Ardian Finance 10k} dataset~\cite{argimi2024Datasets}. This open-source resource comprises $52{,}000$ financial annual reports, written in English and extracted from their original PDF format, covering the period from the late 90s to 2024.
For our evaluation, we focused on a specific 81-page report from this corpus, from which we manually constructed $50$ question/answer/source-page triplets. Answers were generated both with and without the retrieval of the relevant context pages. The task was framed as detecting ``missing context'' hallucinations—instances where the model hallucinates facts due to the specific absence of the retrieved information—based solely on the entropic signature of the answer.

\subsection{Evaluation Protocol}
\subsubsection{Metrics}
The primary metrics used to evaluate the performance of our hallucination detection methods are the Area Under the Precision-Recall Curve (PR-AUC) and the Area Under the Receiver Operating Characteristic Curve (ROC-AUC). As both metrics led to identical conclusions, we report only ROC-AUC scores for conciseness.

\begin{table*}[t]
\centering
\begin{tabular}{@{}lm{3cm} cccc@{}}
    \toprule
    \multirow{2}{*}{\textbf{Dataset}} & \multirow{2}{*}{\textnormal{\textbf{Model}}} & \multicolumn{2}{c}{\textbf{\textit{Unsupervised}}} & \multicolumn{2}{c}{\textbf{\textit{Supervised}}} \\
    \cmidrule(lr){3-4} \cmidrule(lr){5-6}
    & & SCGPT \cite{manakulSelfCheckGPTZeroResourceBlackBox2023} & \begin{tabular}{@{}c@{}}EPR \\ (ours)\end{tabular} & HalluDetect \cite{quevedo2024detecting} & \begin{tabular}{@{}c@{}}WEPR \\ (ours)\end{tabular} \\
    \midrule
    \multirow{4}{*}{\begin{tabular}{@{}l@{}}
    TriviaQA~\cite{joshiTriviaQALargeScale2017}\\(Hallucination\\Detection)
    \end{tabular}}
    & \texttt{Mistral-Small-24B} & 79.0 & 74.6 & 78.7 & \textbf{82.0} \\
    & \texttt{Falcon-3-10B}      & 70.1 & 75.4 & 79.0 & \textbf{84.1} \\
    & \texttt{Phi-4 (14.7B)}     & 71.4 & 78.2 & 83.8 & \textbf{85.4} \\
    & \texttt{Ministral-8B-2410} & 81.1 & 81.4 & \textbf{86.1} & 85.8 \\
    \midrule
    \multirow{4}{*}{\begin{tabular}{@{}l@{}} WebQuestions~\cite{berantSemanticParsingFreebase2013}\\ (Hallucination\\Detection\\Generalization)\end{tabular}}
    & \texttt{Mistral-Small-24B} & 59.3 & 62.5 & 62.8 & \textbf{64.8} \\
    & \texttt{Falcon-3-10B}      & 65.8 & 68.2 & 69.3 & \textbf{73.2} \\
    & \texttt{Phi-4} (14.7B)     & 65.0 & 65.2 & 66.3 & \textbf{66.6} \\
    & \texttt{Ministral-8B-2410} & 66.2 & 65.4 & 71.6 & \textbf{72.6} \\
    \midrule
    \multirow{4}{*}{\begin{tabular}{@{}l@{}} ArGiMi-Ardian~\cite{argimi2024Datasets} \\ (Missing Context \\ Detection)\end{tabular}}
    & \texttt{Mistral-Small-24B} & 64.2 & 81.0 & 81.1 & \textbf{82.8} \\
    & \texttt{Falcon-3-10B}      & 67.6 & 82.8 & 87.0 & \textbf{89.3} \\
    & \texttt{Phi-4 (14.7B)}     & 79.7 & 91.0 & 88.5 & \textbf{93.6} \\
    & \texttt{Ministral-8B-2410} & 65.9 & 73.1 & 81.6 & \textbf{81.8} \\
    \bottomrule
    \vspace{0.1cm}
\end{tabular}
  \caption{ROC-AUC of compared methods across different LLMs, using $K=15$ accessible log-probabilities, on: (i) hallucination detection on TriviaQA; (ii) generalization to WebQuestions (supervised models trained on TriviaQA); and (iii) missing context detection on ArGiMi-Ardian (also trained on TriviaQA).}\label{tab:combined_metrics}
\end{table*}

\subsubsection{Comparison to existing methods}
We benchmark our proposed methods, EPR (unsupervised) and WEPR (supervised), against two state-of-the-art black-box detection techniques:
\begin{itemize}
    \item SelfCheckGPT~\cite{manakulSelfCheckGPTZeroResourceBlackBox2023}: A multi-shot method that measures the consistency across several answers generated for the same query. We used the BertScore variant with $10$ generated samples.
    \item HalluDetect~\cite{quevedo2024detecting}: A single-shot method that, similar to our approach, engineers features from model probabilities to train a detector.
\end{itemize}
This selection allows for a direct comparison with a powerful multi-shot approach (SelfCheckGPT) and a closely related single-shot competitor (HalluDetect), for which we used the same number of accessible log-probabilities (K=15) to ensure a fair comparison.

\subsubsection{Training and Evaluation Strategy}
The supervised logistic regression model have been trained only on TriviaQA data (generated sequences and their labels), as being the only dataset with annotation.
For evaluating these methods on TriviaQA, we generated training and testing splits by grouping the data points by their original query. This process ensures that no data points originating from the same query appear in the training and testing sets for a given fold, mitigating potential data leakage and overestimating generalization performance. To ensure stability and robust performance estimation, particularly given potential variability in smaller datasets or specific model-dataset pairings, we employed bootstrapping on $1000$ iterations during model evaluation.

\subsection{Results}\label{sec:results}
\subsubsection{Hallucination Classification Performance}

Table~\ref{tab:combined_metrics} shows results on TriviaQA for hallucination detection. The EPR baseline demonstrates notable discriminative ability, with ROC-AUC scores ranging from $74.6$ (\texttt{Mistral-Small-3.1-24B}) to $81.4$ (\texttt{Ministral-8B-Instruct-2410}). For a similar computational overhead as other single-shot methods (around $80\pm 20~\unit{\micro\second}$ per score) WEPR consistently matches or outperforms EPR, SelfCheckGPT, and HalluDetect across all LLMs. Note that multi-shot methods are multiple orders of magnitude more computationally costly (at least $10 \unit{\second}$ for SelfCheckGPT).
We also verified the robustness of our approach to sampling parameters; for instance, on \texttt{Falcon-3-10B}, reducing the temperature to $T_{\text{samp}}=0.6$ resulted in a negligible change in ROC-AUC ($< 1.0$ point difference). Since the log-probabilities used for WEPR features are derived from raw logits, this result confirms that our entropic signal remains highly discriminative even when applied to the more deterministic generation trajectories typical of conservative decoding.
Regarding the generalization on WebQuestions (Table~\ref{tab:combined_metrics}), we observe that although performance is generally lower on this dataset, WEPR maintains its advantage across LLMs.
\begin{figure}
    \centering
    \includegraphics[width=\linewidth]{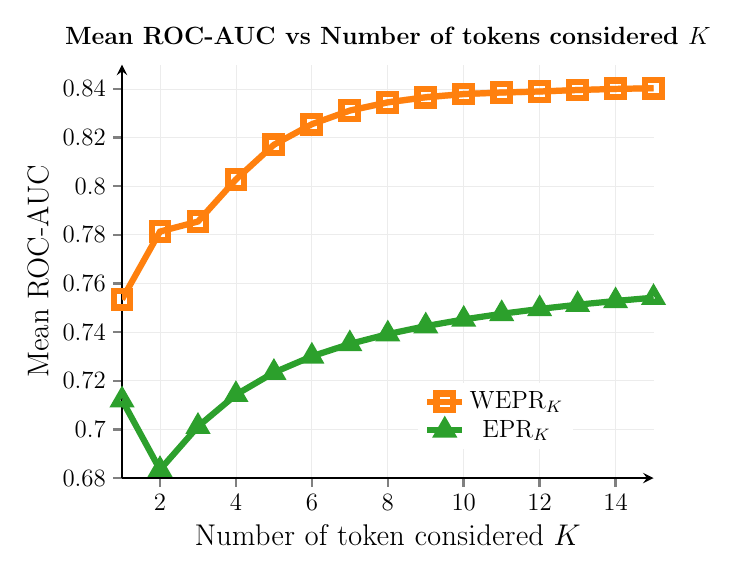}
    \caption{Mean ROC-AUC with respect to $K$ for Falcon-3 on TriviaQA.}
    \Description{There is two curves. The one corresponding to WEPR has its performance increasing with $K$, and seems to reach a plateau around $K=10$---$12$.}
    \label{fig:roc_aucs_with_k}
\end{figure}

\begin{figure}
    \centering
    \includegraphics[width=\linewidth]{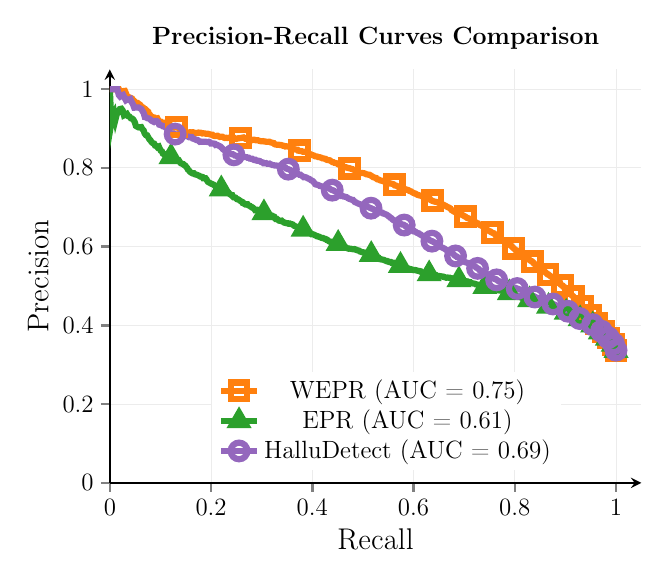}
    \caption{Precision-Recall curves of 3 methods for Falcon-3 on TriviaQA.}
    \Description{A sample PR curve, for the 3 methods. The curve corresponding to WEPR is consistently above the HalluDetect one, which is consistently above the EPR one.}
    \label{fig:PR_curves}
\end{figure}
The improved classification performance of the WEPR model is further illustrated by the sample Precision-Recall curves for \texttt{Falcon-3-10B} on TriviaQA (Figure~\ref{fig:PR_curves}), where the WEPR curve consistently dominates that of the EPR baseline and the HalluDetect \cite{quevedo2024detecting} method.

\begin{table*}
    \centering
    \setlength{\tabcolsep}{10pt}
    \begin{tabular}{@{}l c c c@{}}
        \toprule
        \textbf{Task annotated} & \textbf{LLM-as-a-Judge} & \textbf{Sentence-level} & \textbf{Token-level} \\
        \midrule
        Samples         & 1,333 & 312  & 334  \\
        Agreement (\%)  & 95.7  & 77.8 & 78.1 \\
        \bottomrule
    \end{tabular}
    \caption{Human agreement with (i) LLM annotations validating labels in Section~\ref{sec:results}, and (ii) WEPR scores at sentence and token granularity, measuring adoption potential.}
    \label{tab:annotations}
\end{table*}

\subsubsection{RAG uncertainty}

Table~\ref{tab:combined_metrics} presents the performance of compared methods (trained on the general TriviaQA dataset for supervised ones) for the missing context detection task. The results show that the WEPR model significantly outperforms the other methods in this domain-specific RAG scenario.
This indicates that an entropic detector trained on a general QA dataset can transfer, to a notable extent, to specialized domains for identifying responses generated with insufficient context. Such a mechanism can be readily applied in RAG pipelines to flag outputs where context retrieval may have been inadequate or where the model is otherwise struggling, thereby triggering deeper information retrieval or alerting the user.
\subsubsection{Effect of the Number of Accessible Log-probabilities ($K$)}
We investigated the impact of $K$ on hallucination detection performance. Figure~\ref{fig:roc_aucs_with_k} displays the evolution of the mean ROC-AUC for \texttt{Falcon-3-10B} on TriviaQA as $K$ varies. The results indicate that performance generally improves with increasing ranked token contributions up to a certain point. Notably, for several models, the ROC-AUC tends to plateau when $K$ reaches approximately $8$ to $10$. This suggests that incorporating entropic contributions beyond the top $\sim$10 ranks may yield diminishing returns for hallucination detection. This finding has practical significance, implying that effective detection can be achieved even when API access is limited to a relatively small number of top logprobs, thereby reducing data requirements and potentially API costs.

\subsubsection{Human adoption analysis}
We wanted to measure the perceived downstream quality of our $\mathrm{WEPR}$ scoring methods, at the two possible granularities: sentence-wise ($\sigma\left(\mathrm{WEPR}_{\beta,\gamma}(q,\mathcal{T})\right)$ and token-wise (Eq.~\ref{eq:token_score}).
At least three independent annotators (from the same pool of researchers) rated each sample as acceptable or not (mean $3.76$). The adoption score is then computed based on the majority vote, with results compiled in Table~\ref{tab:annotations}.
The high level of satisfaction for token scores ($78.1 \%$) indicates that the learned coefficients $\beta_k$ from our WEPR model can be used to compute a weighted entropic contribution for each token $j$ in a sequence, thus offering a finer-grained view of uncertainty hotspots within the generated text (see text colors of Figure~\ref{fig:overview}).

\section{Limitations}\label{sec:limitations}

While our methods demonstrate promising results for one-shot, black-box hallucination detection, several limitations warrant acknowledgment. 
First, our experiments were conducted on mid-sized LLMs (approximately 8B–24B parameters). Performance on bigger models (e.g., 100B+ parameters or closed-source APIs like GPT-5) remains to be tested. 
Second, the current study focuses on QA tasks that yield relatively short, factual answers. We did not study tasks requiring extensive multi-step reasoning or lengthy outputs, where the aggregated entropy signal might become less distinct or different types of uncertainty could dominate.
Moreover, the method may produce false positives by flagging model hesitation related to stylistic choices or sentence formulation, as opposed to uncertainty about the factual content of the answer. This effect was observed to be more pronounced at the beginning of generated sequences.
Finally, our approach, like many other uncertainty-based methods, is inherently limited in its ability to detect "high-certainty hallucinations"~\cite{simhiTrustMeIm2025}—instances where the model generates incorrect information with low output entropy (i.e., high confidence) due to memorization errors. If an LLM has strongly learned erroneous facts during its training, entropic measures based on output probabilities may not flag such confident fabrications.

\section{Conclusion}
\label{sec:conclusion}
We addressed one-shot hallucination detection in black-box LLMs by introducing a methodology rooted in information-theoretic principles, leveraging only the top-$K$ token log-probabilities typically exposed by APIs. Our primary contribution is a supervised approach that utilizes features representing individual top-ranked tokens mean and maximal entropic contributions. Experiments across QA datasets and multiple LLMs show that the learned estimator outperforms a baseline Entropy Production Rate (EPR). Notably, strong detection performance is achieved even with as few as $K\leq10$ accessible log-probabilities, underscoring practical efficiency under API constraints. We also showcased utility in a specialized financial setting for RAG and highlighted token-level uncertainty visualization.

This work provides a practical technique for enhancing the trustworthiness of LLM responses in both general QA and RAG from a single generation pass. Future work may extend this approach to more complex reasoning tasks and study the interpretability of the learned entropic weights.

\begin{acks}
    This work was done as part of the ArGiMi project, funded by BPIFrance under the France2030 national effort towards numerical common goods. CM wishes to thank Antoine Delaby for preliminary discussions on RAG applications, along with Gabriel Trier, Lucas Tramonte and Rayane Bouaita for their help on the question dataset. HR also wishes to thank Yves-Roland Kouayip and Mathieu Crochet from Artefact for their help in creating the annotation tool.
\end{acks}

\bibliographystyle{unsrt}

\begin{thebibliography}{10}
	
	\bibitem{petroniLanguageModelsKnowledge2019}
	Fabio Petroni, Tim Rockt{\"a}schel, Sebastian Riedel, Patrick Lewis, Anton
	Bakhtin, Yuxiang Wu, and Alexander Miller.
	\newblock Language {{Models}} as {{Knowledge Bases}}?
	\newblock In {\em Proceedings of the 2019 {{Conference}} on {{Empirical
				Methods}} in {{Natural Language Processing}} and the 9th {{International
				Joint Conference}} on {{Natural Language Processing}} ({{EMNLP-IJCNLP}})},
	pages 2463--2473, Hong Kong, China, 2019. Association for Computational
	Linguistics.
	
	\bibitem{weiEmergentAbilitiesLarge2022}
	Jason Wei, Yi~Tay, Rishi Bommasani, Colin Raffel, Barret Zoph, Sebastian
	Borgeaud, Dani Yogatama, Maarten Bosma, Denny Zhou, Donald Metzler, Ed~H.
	Chi, Tatsunori Hashimoto, Oriol Vinyals, Percy Liang, Jeff Dean, and William
	Fedus.
	\newblock Emergent {{Abilities}} of {{Large Language Models}}, October 2022.
	
	\bibitem{openaiGPT4TechnicalReport2024}
	OpenAI et. al.
	\newblock {{GPT-4 Technical Report}}, March 2024.
	
	\bibitem{rohrbachObjectHallucinationImage2019}
	Anna Rohrbach, Lisa~Anne Hendricks, Kaylee Burns, Trevor Darrell, and Kate
	Saenko.
	\newblock Object {{Hallucination}} in {{Image Captioning}}, March 2019.
	
	\bibitem{bakerHallucinatingFaces2000}
	S.~Baker and T.~Kanade.
	\newblock Hallucinating faces.
	\newblock In {\em Proceedings {{Fourth IEEE International Conference}} on
		{{Automatic Face}} and {{Gesture Recognition}} ({{Cat}}. {{No}}.
		{{PR00580}})}, pages 83--88, March 2000.
	
	\bibitem{jiSurveyHallucinationNatural2023a}
	Ziwei Ji, Nayeon Lee, Rita Frieske, Tiezheng Yu, Dan Su, Yan Xu, Etsuko Ishii,
	Yejin Bang, Delong Chen, Wenliang Dai, Ho~Shu Chan, Andrea Madotto, and
	Pascale Fung.
	\newblock Survey of {{Hallucination}} in {{Natural Language Generation}}.
	\newblock {\em ACM Computing Surveys}, 55(12):1--38, December 2023.
	
	\bibitem{sriramananLLMCheckInvestigatingDetection}
	Gaurang Sriramanan, Siddhant Bharti, Vinu~Sankar Sadasivan, Shoumik Saha,
	Priyatham Kattakinda, and Soheil Feizi.
	\newblock Llm-check: Investigating detection of hallucinations in large
	language models.
	\newblock In A.~Globerson, L.~Mackey, D.~Belgrave, A.~Fan, U.~Paquet,
	J.~Tomczak, and C.~Zhang, editors, {\em Advances in Neural Information
		Processing Systems}, volume~37, pages 34188--34216. Curran Associates, Inc.,
	2024.
	
	\bibitem{kimMedicalHallucinationsFoundation2025}
	Yubin Kim, Hyewon Jeong, Shan Chen, Shuyue~Stella Li, Chanwoo Park, Mingyu Lu,
	Kumail Alhamoud, Jimin Mun, Cristina Grau, Minseok Jung, et~al.
	\newblock Medical hallucinations in foundation models and their impact on
	healthcare.
	\newblock {\em arXiv preprint arXiv:2503.05777}, 2025.
	
	\bibitem{liuExploringEvaluatingHallucinations2024}
	Fang Liu, Yang Liu, Lin Shi, Houkun Huang, Ruifeng Wang, Zhen Yang, Li~Zhang,
	Zhongqi Li, and Yuchi Ma.
	\newblock Exploring and {{Evaluating Hallucinations}} in {{LLM-Powered Code
			Generation}}, May 2024.
	
	\bibitem{sullivanIntroductionUncertaintyQuantification2015}
	T.~J. Sullivan.
	\newblock {\em Introduction to {{Uncertainty Quantification}}}.
	\newblock Springer, December 2015.
	
	\bibitem{manakulSelfCheckGPTZeroResourceBlackBox2023}
	Potsawee Manakul, Adian Liusie, and Mark J.~F. Gales.
	\newblock {{SelfCheckGPT}}: {{Zero-Resource Black-Box Hallucination Detection}}
	for {{Generative Large Language Models}}, October 2023.
	
	\bibitem{liuUncertaintyQuantificationConfidence2025}
	Xiaoou Liu, Tiejin Chen, Longchao Da, Chacha Chen, Zhen Lin, and Hua Wei.
	\newblock Uncertainty {{Quantification}} and {{Confidence Calibration}} in
	{{Large Language Models}}: {{A Survey}}, March 2025.
	
	\bibitem{kadavathLanguageModelsMostly2022}
	Saurav Kadavath, Tom Conerly, Amanda Askell, Tom Henighan, Dawn Drain, Ethan
	Perez, Nicholas Schiefer, Zac Hatfield-Dodds, Nova DasSarma, Eli
	Tran-Johnson, et~al.
	\newblock Language models (mostly) know what they know.
	\newblock {\em arXiv preprint arXiv:2207.05221}, 2022.
	
	\bibitem{azariaInternalStateLLM2023}
	Amos Azaria and Tom Mitchell.
	\newblock The {{Internal State}} of an {{LLM Knows When It}}'s {{Lying}},
	October 2023.
	
	\bibitem{gal2016uncertainty}
	Yarin Gal.
	\newblock {\em Uncertainty in Deep Learning}.
	\newblock PhD thesis, University of Cambridge, 2016.
	
	\bibitem{hullermeier2021aleatoric}
	Eyke H{\"u}llermeier and Willem Waegeman.
	\newblock Aleatoric and epistemic uncertainty in machine learning: {{An}}
	introduction to concepts and methods.
	\newblock {\em Machine learning}, 110(3):457--506, 2021.
	
	\bibitem{postels2019sampling}
	Janis Postels, Francesco Ferroni, Huseyin Coskun, Nassir Navab, and Federico
	Tombari.
	\newblock Sampling-free epistemic uncertainty estimation using approximated
	variance propagation.
	\newblock In {\em Proceedings of the {{IEEE}}/{{CVF}} International Conference
		on Computer Vision}, pages 2931--2940, 2019.
	
	\bibitem{gal2016dropout}
	Yarin Gal and Zoubin Ghahramani.
	\newblock Dropout as a bayesian approximation: {{Representing}} model
	uncertainty in deep learning.
	\newblock In {\em International Conference on Machine Learning}, pages
	1050--1059. PMLR, 2016.
	
	\bibitem{quevedo2024detecting}
	Ernesto Quevedo, Jorge~Yero Salazar, Rachel Koerner, Pablo Rivas, and Tomas
	Cerny.
	\newblock Detecting hallucinations in large language model generation: A token
	probability approach.
	\newblock In {\em World Congress in Computer Science, Computer Engineering \&
		Applied Computing}, pages 154--173. Springer, 2024.
	
	\bibitem{shannonMathematicalTheoryCommunication1948}
	C.~E. Shannon.
	\newblock A mathematical theory of communication.
	\newblock {\em The Bell System Technical Journal}, 27(3):379--423, July 1948.
	
	\bibitem{gabrieEntropyMutualInformation2018}
	Marylou Gabri{\'e}, Andre Manoel, Cl{\'e}ment Luneau, jean {barbier}, Nicolas
	Macris, Florent Krzakala, and Lenka Zdeborov{\'a}.
	\newblock Entropy and mutual information in models of deep neural networks.
	\newblock In {\em Advances in {{Neural Information Processing Systems}}},
	volume~31. Curran Associates, Inc., 2018.
	
	\bibitem{kweonUncertaintyQuantificationDecomposition2025}
	Wonbin Kweon, Sanghwan Jang, SeongKu Kang, and Hwanjo Yu.
	\newblock Uncertainty {{Quantification}} and {{Decomposition}} for {{LLM-based
			Recommendation}}.
	\newblock In {\em Proceedings of the {{ACM}} on {{Web Conference}} 2025}, pages
	4889--4901, April 2025.
	
	\bibitem{kuhnSemanticUncertaintyLinguistic2023}
	Lorenz Kuhn, Yarin Gal, and Sebastian Farquhar.
	\newblock Semantic {{Uncertainty}}: {{Linguistic Invariances}} for
	{{Uncertainty Estimation}} in {{Natural Language Generation}}, April 2023.
	
	\bibitem{farquharDetectingHallucinationsLarge2024}
	Sebastian Farquhar, Jannik Kossen, Lorenz Kuhn, and Yarin Gal.
	\newblock Detecting hallucinations in large language models using semantic
	entropy.
	\newblock {\em Nature}, 630(8017):625--630, June 2024.
	
	\bibitem{cecereMonteCarloTemperature2025}
	Nicola Cecere, Andrea Bacciu, Ignacio~Fern{\'a}ndez Tob{\'i}as, and Amin
	Mantrach.
	\newblock Monte {{Carlo Temperature}}: A robust sampling strategy for {{LLM}}'s
	uncertainty quantification methods, February 2025.
	
	\bibitem{liuEnhancingHallucinationDetection2025}
	Litian Liu, Reza Pourreza, Sunny Panchal, Apratim Bhattacharyya, Yao Qin, and
	Roland Memisevic.
	\newblock Enhancing {{Hallucination Detection}} through {{Noise Injection}},
	February 2025.
	
	\bibitem{jooEntropybasedSentencelevelHallucination2025}
	Eojin Joo, Young-Jun Lee, and Ho-Jin Choi.
	\newblock Entropy-based {{Sentence-level Hallucination Score}} in {{Large
			Language Models}}.
	\newblock In {\em 2025 {{IEEE International Conference}} on {{Big Data}} and
		{{Smart Computing}} ({{BigComp}})}, pages 77--78, February 2025.
	
	\bibitem{kossenSemanticEntropyProbes2024}
	Jannik Kossen, Jiatong Han, Muhammed Razzak, Lisa Schut, Shreshth Malik, and
	Yarin Gal.
	\newblock Semantic {{Entropy Probes}}: {{Robust}} and {{Cheap Hallucination
			Detection}} in {{LLMs}}, June 2024.
	
	\bibitem{duanShiftingAttentionRelevance2024}
	Jinhao Duan, Hao Cheng, Shiqi Wang, Alex Zavalny, Chenan Wang, Renjing Xu,
	Bhavya Kailkhura, and Kaidi Xu.
	\newblock Shifting {{Attention}} to {{Relevance}}: {{Towards}} the {{Predictive
			Uncertainty Quantification}} of {{Free-Form Large Language Models}}, May
	2024.
	
	\bibitem{simhiTrustMeIm2025}
	Adi Simhi, Itay Itzhak, Fazl Barez, Gabriel Stanovsky, and Yonatan Belinkov.
	\newblock Trust {{Me}}, {{I}}'m {{Wrong}}: {{High-Certainty Hallucinations}} in
	{{LLMs}}, February 2025.
	
	\bibitem{maEstimatingLLMUncertainty2025}
	Huan Ma, Jingdong Chen, Joey~Tianyi Zhou, Guangyu Wang, and Changqing Zhang.
	\newblock Estimating {{LLM Uncertainty}} with {{Evidence}}, May 2025.
	
	\bibitem{brillouinScienceInformationTheory2013}
	Leon Brillouin.
	\newblock {\em Science and {{Information Theory}}}.
	\newblock Courier Corporation, July 2013.
	
	\bibitem{mezardInformationPhysicsComputation2008}
	Marc Mezard and Andrea Montanari.
	\newblock {\em Information, physics, and computation}.
	\newblock Oxford University Press, 2009.
	
	\bibitem{MistralSmall31}
	Mistral {{Small}} 3.1 {\textbar} {{Mistral AI}}.
	\newblock https://mistral.ai/news/mistral-small-3-1.
	
	\bibitem{Falcon3}
	TII Falcon-LLM Team.
	\newblock The falcon 3 family of open models.
	\newblock https://huggingface.co/blog/falcon3, December 2024.
	
	\bibitem{abdinPhi4TechnicalReport2024}
	Marah Abdin, Jyoti Aneja, Harkirat Behl, S{\'e}bastien Bubeck, Ronen Eldan,
	Suriya Gunasekar, Michael Harrison, Russell~J Hewett, Mojan Javaheripi, Piero
	Kauffmann, et~al.
	\newblock Phi-4 technical report.
	\newblock {\em arXiv preprint arXiv:2412.08905}, 2024.
	
	\bibitem{MinistralMinistrauxMistral}
	{Un Ministral, des Ministraux {\textbar} Mistral AI}.
	\newblock https://mistral.ai/fr/news/ministraux.
	
	\bibitem{joshiTriviaQALargeScale2017}
	Mandar Joshi, Eunsol Choi, Daniel~S. Weld, and Luke Zettlemoyer.
	\newblock {{TriviaQA}}: {{A Large Scale Distantly Supervised Challenge
			Dataset}} for {{Reading Comprehension}}, May 2017.
	
	\bibitem{berantSemanticParsingFreebase2013}
	Jonathan Berant, Andrew Chou, Roy Frostig, and Percy Liang.
	\newblock Semantic {{Parsing}} on {{Freebase}} from {{Question-Answer Pairs}}.
	\newblock In David Yarowsky, Timothy Baldwin, Anna Korhonen, Karen Livescu, and
	Steven Bethard, editors, {\em Proceedings of the 2013 {{Conference}} on
		{{Empirical Methods}} in {{Natural Language Processing}}}, pages 1533--1544,
	Seattle, Washington, USA, October 2013. Association for Computational
	Linguistics.
	
	\bibitem{lewisRetrievalAugmentedGenerationKnowledgeIntensive2021}
	Patrick Lewis, Ethan Perez, Aleksandra Piktus, Fabio Petroni, Vladimir
	Karpukhin, Naman Goyal, Heinrich K{\"u}ttler, Mike Lewis, Wen-tau Yih, Tim
	Rockt{\"a}schel, Sebastian Riedel, and Douwe Kiela.
	\newblock Retrieval-{{Augmented Generation}} for {{Knowledge-Intensive NLP
			Tasks}}, April 2021.
	
	\bibitem{argimi2024Datasets}
	Hicham Randrianarivo, Charles Moslonka, Arthur Garnier, and Emmanuel Malherbe.
	\newblock The {{ArGiMi}} datasets.
	\newblock
	https://huggingface.co/datasets/artefactory/Argimi-Ardian-Finance-10k-text,
	2024.
	
	\bibitem{kwon2023efficient}
	Woosuk Kwon, Zhuohan Li, Siyuan Zhuang, Ying Sheng, Lianmin Zheng, Cody~Hao Yu,
	Joseph~E. Gonzalez, Hao Zhang, and Ion Stoica.
	\newblock Efficient memory management for large language model serving with
	{{PagedAttention}}.
	\newblock In {\em Proceedings of the {{ACM SIGOPS}} 29th Symposium on Operating
		Systems Principles}, 2023.
	
	\bibitem{zhengJudgingLLMJudgeMTBench2023}
	Lianmin Zheng, Wei-Lin Chiang, Ying Sheng, Siyuan Zhuang, Zhanghao Wu, Yonghao
	Zhuang, Zi~Lin, Zhuohan Li, Dacheng Li, Eric~P. Xing, Hao Zhang, Joseph~E.
	Gonzalez, and Ion Stoica.
	\newblock Judging {{LLM-as-a-Judge}} with {{MT-Bench}} and {{Chatbot Arena}},
	December 2023.
	
	\bibitem{wangLargeLanguageModels2023}
	Peiyi Wang, Lei Li, Liang Chen, Zefan Cai, Dawei Zhu, Binghuai Lin, Yunbo Cao,
	Qi~Liu, Tianyu Liu, and Zhifang Sui.
	\newblock Large {{Language Models}} are not {{Fair Evaluators}}, August 2023.
	
	\bibitem{adlakhaEvaluatingCorrectnessFaithfulness2024}
	Vaibhav Adlakha, Parishad BehnamGhader, Xing~Han Lu, Nicholas Meade, and Siva
	Reddy.
	\newblock Evaluating {{Correctness}} and {{Faithfulness}} of
	{{Instruction-Following Models}} for {{Question Answering}}.
	\newblock {\em Transactions of the Association for Computational Linguistics},
	12:681--699, May 2024.
	
	\bibitem{rahmaniLLMJudgeLLMsRelevance2024}
	Hossein~A. Rahmani, Emine Yilmaz, Nick Craswell, Bhaskar Mitra, Paul Thomas,
	Charles L.~A. Clarke, Mohammad Aliannejadi, Clemencia Siro, and Guglielmo
	Faggioli.
	\newblock {{LLMJudge}}: {{LLMs}} for {{Relevance Judgments}}, August 2024.
	
	\bibitem{teamGemma3Technical2025}
	Gemma et.~al. Team.
	\newblock Gemma 3 {{Technical Report}}, March 2025.
	
\end{thebibliography}

\end{document}